\documentclass[10pt,journal,compsoc]{IEEEtran}

\usepackage{amsmath,amssymb}
\usepackage{algorithm}
\usepackage{algorithmicx}
\usepackage{graphics}
\usepackage{cite}
\usepackage{color}
\usepackage{ragged2e}
\usepackage{graphicx}
\usepackage[normalem]{ulem}
\usepackage{overpic}
\usepackage{multirow}
\usepackage{makecell}
\usepackage{colortbl}
\usepackage[breaklinks=true,colorlinks]{hyperref}
\usepackage{pifont} %

\usepackage{silence}
\graphicspath{{./Imgs/},{./Imgs/authors/}}
\DeclareGraphicsExtensions{.pdf,.jpg,.png}

\newcommand{\figref}[1]{Fig.~\ref{#1}}
\newcommand{\tabref}[1]{Table~\ref{#1}}
\newcommand{\equref}[1]{Equ.~\ref{#1}}
\newcommand{\secref}[1]{$\S$\ref{#1}}

\ifdefined \GramaCheck
  \newcommand{\CheckRmv}[1]{}
  \renewcommand{\equref}[1]{Equation 1}
  \renewcommand{\figref}[1]{Figure 1}
  \renewcommand{\tabref}[1]{Table 1}
\else
  \newcommand{\CheckRmv}[1]{#1}
\fi

\newcommand{\tabincell}[2]{\begin{tabular}{@{}#1@{}}#2\end{tabular}}
\newcommand{\parahead}[1]{\noindent\textbf{#1}\quad}
\newcommand{\myPara}[1]{\vspace{.12in}\noindent\textbf{#1}\quad}

\definecolor{mygray}{gray}{.92}

\def\ie{\emph{i.e.}}
\def\eg{\emph{e.g.}}

\def\etal{{\em et al.~}}

\begin{document}

\title{P2T: Pyramid Pooling Transformer \\ for Scene Understanding}

\markboth{IEEE TRANSACTIONS ON PATTERN ANALYSIS AND MACHINE INTELLIGENCE}%
{W\MakeLowercase{u} \MakeLowercase{\textit{et al.}}: P2T: Pyramid Pooling Transformer for Scene Understanding}

\author{
  Yu-Huan Wu, Yun Liu, Xin Zhan, and Ming-Ming Cheng
  \IEEEcompsocitemizethanks{%
    \IEEEcompsocthanksitem Y.-H.~Wu and M.-M.~Cheng are with TMCC, 
    College of Computer Science, Nankai University, Tianjin, China. 
    (E-mail: wuyuhuan@mail.nankai.edu.cn, cmm@nankai.edu.cn)
    \IEEEcompsocthanksitem Y. Liu is with Institute for 
    Infocomm Research (I2R), Agency for Science, Technology 
    and Research (A*STAR), Singapore.
    (E-mail: vagrantlyun@gmail.com)
    \IEEEcompsocthanksitem X. Zhan is with Alibaba DAMO Academy, Hangzhou, China. 
    \IEEEcompsocthanksitem The first two authors contributed equally to this work.
    \IEEEcompsocthanksitem Corresponding author: M.-M. Cheng. (E-mail: cmm@nankai.edu.cn)
    \IEEEcompsocthanksitem This work is done while Y.-H. Wu is a research intern at Alibaba DAMO Academy.
  }
}

\IEEEtitleabstractindextext{%
\begin{abstract} \justifying
Recently, the vision transformer has achieved great success by pushing the state-of-the-art of various vision tasks. One of the most challenging problems in the vision transformer is that the large sequence length of image tokens leads to high computational cost (quadratic complexity). 
A popular solution to this problem is to use a single pooling operation to reduce the sequence length. This paper considers how to improve existing vision transformers, where the pooled feature extracted by a single pooling operation seems less powerful. 
To this end, we note that pyramid pooling has been demonstrated to be effective in various vision tasks owing to its powerful ability in context abstraction. However, pyramid pooling has not been explored in backbone network design. 
To bridge this gap, we propose to adapt pyramid pooling to Multi-Head Self-Attention (MHSA) in the vision transformer, simultaneously reducing the sequence length and capturing powerful contextual features.
Plugged with our pooling-based MHSA, we build a universal vision transformer backbone, dubbed Pyramid Pooling Transformer (P2T). Extensive experiments demonstrate that, 
when applied P2T as the backbone network, 
it shows substantial superiority in various vision tasks such as image classification, semantic segmentation, object detection, and instance segmentation, compared to previous CNN- and transformer-based networks. The code will be released at \url{https://github.com/yuhuan-wu/P2T}.
\end{abstract}
}

\maketitle

\IEEEdisplaynontitleabstractindextext
\IEEEpeerreviewmaketitle

\IEEEraisesectionheading{\section{Introduction}}\label{sec:intro}
\IEEEPARstart{I}{n} the past decade, convolutional neural networks (CNNs) 
have dominated computer vision and achieved many great stories
\cite{krizhevsky2012imagenet,simonyan2014very,szegedy2015going,he2016deep,huang2017densely,hu2020squeeze,tan2019efficientnet,wu2021mobilesal,gao2021res2net}.
The state-of-the-art of various vision tasks on many large-scale datasets
has been significantly pushed forward
\cite{russakovsky2015imagenet,lin2014microsoft,everingham2010pascal,cordts2016cityscapes,zhou2017scene}.
In an orthogonal field, \ie, natural language processing (NLP), 
the dominating technique is transformer \cite{vaswani2017attention}.
Transformer entirely relies on self-attention to capture the long-range 
global relationships and has achieved brilliant successes.
Considering that global information is also essential for vision tasks,
a proper adaption of the transformer \cite{vaswani2017attention} should be 
useful to overcome the limitation of CNNs, \ie, CNNs usually enlarge the
receptive field by stacking more layers.

Lots of efforts are dedicated to exploring such a proper adaption of the
transformer \cite{vaswani2017attention}.
Some early attempts use CNNs \cite{simonyan2014very,he2016deep} to extract 
deep features that are fed into transformers for further processing
and regressing the targets 
\cite{carion2020end,zhu2020deformable,hu2021transformer}.
Dosovitskiy \etal \cite{dosovitskiy2021image} made a thorough success
by applying a pure transformer network for image classification.
They split an image into patches and took each patch as a word/token 
in an NLP application so that transformer can then be directly adopted.
This simple method attains competitive performance on ImageNet
\cite{russakovsky2015imagenet}.
Therefore, a new concept of the vision transformer appears.
In a very short period, a large amount of literature has emerged to
improve the vision transformer \cite{dosovitskiy2021image},
and much better performance than CNNs has been achieved
\cite{heo2021rethinking,wang2021pyramid,liu2021swin,fan2021multiscale,liu2021transformer,liang2021swinir}.

\CheckRmv{
\begin{figure}[t]
    \centering
    \includegraphics[width=\columnwidth]{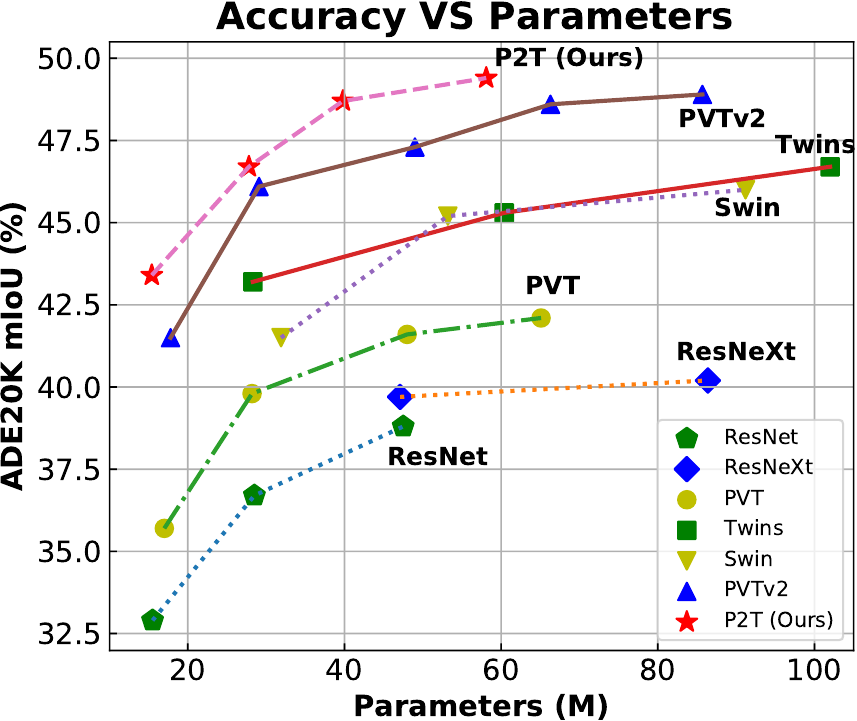}
    \caption{\textbf{Experimental results for semantic segmentation 
    on the ADE20K dataset \cite{zhou2017scene}.}
    Following PVT \cite{wang2021pyramid}, Semantic FPN 
    \cite{kirillov2019panoptic} is chosen as the basic method,
    equipped with different backbone networks, including
    ResNet \cite{he2016deep}, ResNeXt \cite{xie2017aggregated}, PVT \cite{wang2021pyramid}, Twins \cite{chu2021twins}, Swin Transformer \cite{liu2021swin}, PVTv2 \cite{wang2021pvtv2},
    and our P2T.}
    \label{fig:banner}
\end{figure}
}

Nevertheless, there is still one challenging problem in vision transformers, \ie, the length of the data sequence.
When viewing image patches as tokens, the sequence length is much 
longer than in NLP applications.
For example, in NLP, the well-known WMT2014 English-German dataset \cite{bojar2014findings} has 50M English words with 2M sentences, with an average sequence length of 25. 
In contrast, in computer vision, we usually use the image resolution of $224\times 224$ for image classification on the ImageNet dataset \cite{russakovsky2015imagenet}, resulting in the sequence length of 3136 if we use the common patch size of $4\times 4$.
Since the computational and space complexity of 
Multi-Head Self-Attention (MHSA) in the transformer is \textit{quadratic} 
(rather than \textit{linear} in CNNs) to the image size,
directly applying the transformer to vision tasks has a high 
requirement for computational resources.
To make a pure transformer network possible for image classification,
ViT \cite{dosovitskiy2021image} uses large 
patch sizes, \eg, 16 and 32, to reduce the sequence length, and achieves great success for image classification.
Later, many transformer works significantly improve the performance of ViT \cite{dosovitskiy2021image} by introducing the pyramid structure \cite{wang2021pyramid,liu2021swin,fan2021multiscale,liu2021transformer,wu2021cvt,chu2021twins,heo2021rethinking,graham2021levit}, 
where the input layer first uses the small patch size of $4\times 4$ and the sequence length is gradually reduced by merging adjacent image patches.

To further reduce the computational cost of MHSA, PVT \cite{wang2021pyramid} and  MViT \cite{fan2021multiscale} use a single pooling operation to downsample the feature map in the computation of MHSA.
With the pooled features, they model token-to-region relationship rather than 
the expected token-to-token relationship.
Swin Transformer \cite{liu2021swin} proposes to compute MHSA 
within small windows rather than across the whole input, 
modeling local relationships.
It uses a window shift strategy to gradually enlarge the receptive 
field, like CNNs, enlarging the receptive field through 
stacking more layers \cite{han2021demystifying}.
However, an essential characteristic of the vision transformer 
is its \textit{direct global relationship modeling}, which is also why we transfer from CNNs to transformers.

Here, we consider how to improve PVT \cite{wang2021pyramid} and MViT \cite{fan2021multiscale} where the pooled feature extracted by a single pooling operation seems less powerful.
If we can squeeze the input feature into a powerful representation with a short sequence length, we may achieve better performance.
To this end, we note that pyramid pooling
\cite{grauman2005pyramid, lazebnik2006beyond,he2015spatial,zhao2017pyramid} is a long-history technique for computer vision, which extracts contextual information and utilizes multiple pooling operations with different receptive fields and strides onto the input feature map. 
This simple technique has been demonstrated to be effective in various downstream vision tasks such as
semantic segmentation \cite{zhao2017pyramid} and
object detection \cite{he2015spatial}.
Nevertheless, recent pyramid pooling approaches highly rely on a pretrained CNN backbone,
and thus they are limited to a specific task.
In another word, the pyramid pooling technique \textit{\textbf{has not been}} explored in the backbone network design that has broad applications.
Motivated by this, we bridge this gap by adapting pyramid pooling to the vision transformer block for \textit{\textbf{simultaneously}} reducing the sequence length and learning powerful contextual representations.
Pyramid pooling is also very efficient and thus will only induce negligible computational cost for the vision transformer.

We achieve this goal by proposing a new transformer backbone network, \ie, \textbf{Pyramid Pooling Transformer (P2T)}.
We adapt the idea of pyramid pooling to the computation
of multi-head self-attention (MHSA) in the vision transformer, reducing the computational
cost of MHSA and capturing rich contextual information simultaneously.
By applying the new pooling-based MHSA, P2T exhibits stronger ability in feature representation learning
and visual recognition than PVT \cite{wang2021pyramid} and MViT \cite{fan2021multiscale} which are based on a single pooling operation.
We evaluate P2T for various typical vision tasks, like
image classification, semantic segmentation, object detection, and instance segmentation.
Extensive experiments demonstrate that P2T performs better than all previous CNN- and transformer-based backbone networks for these fundamental vision tasks (see \figref{fig:banner} for comparisons on semantic segmentation).

In summary, our main contributions include:
\begin{itemize}
\item We encapsulate pyramid pooling to MHSA, simultaneously reducing the sequence length of image tokens and extracting powerful contextual features.
\item We plug our pooling-based MHSA into the vision transformer to build a new backbone network, \ie, P2T,
making it flexible and powerful for visual recognition.
\item We conduct extensive experiments to demonstrate that, when applied as a backbone network for various scene understanding tasks, P2T achieves substantially better performance than previous CNN- and transformer-based networks.
\end{itemize}

\CheckRmv{
\begin{figure*}[!t]
    \centering
    \includegraphics[width=0.98\textwidth]{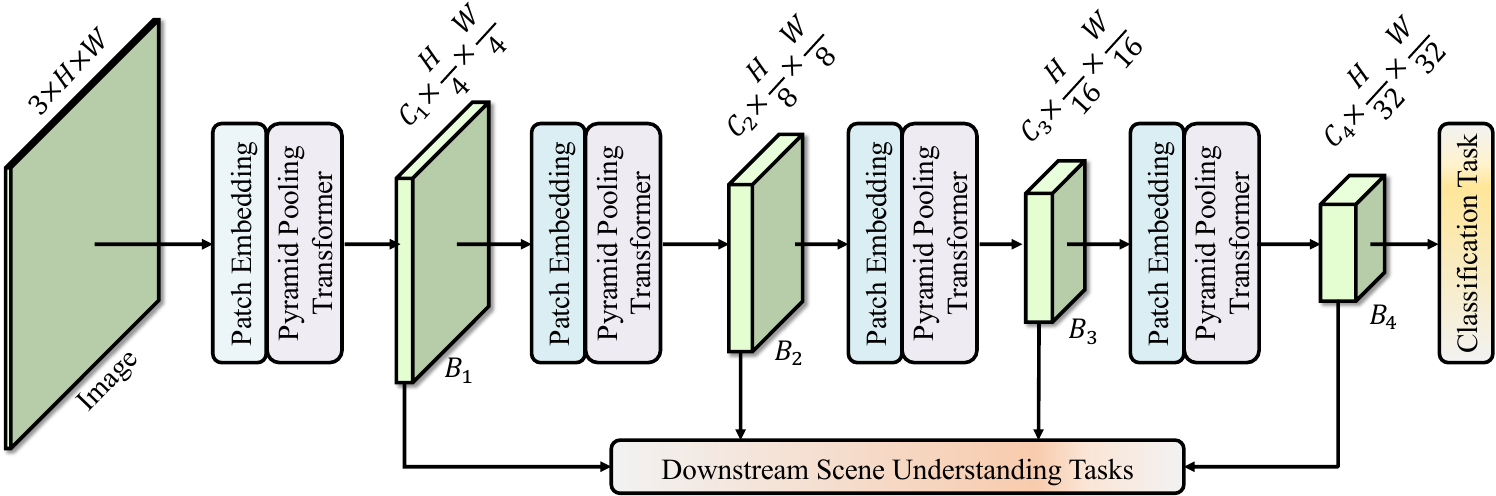}
    \caption{\textbf{Architecture of the proposed P2T network.}
    We replace the traditional MHSA with our pooling-based MHSA.
    The feature maps $\{\mathbf{B}_1, \mathbf{B}_2, \mathbf{B}_3, \mathbf{B}_4\}$ 
    can be used for downstream scene understanding tasks.}
    \label{fig:pipeline}
    \vspace{-.2mm}
\end{figure*}
}

\section{Related Work} 
\subsection{Convolutional Neural Networks}
Since AlexNet \cite{krizhevsky2012imagenet} won the champion 
in the ILSVRC-2012 competition \cite{russakovsky2015imagenet},
numerous advanced techniques have been invented for improving CNNs,
achieving many successful stories in computer vision.
VGG \cite{simonyan2014very} and GoogleNet \cite{szegedy2015going}
first try to deepen CNNs for better image recognition.
Then, ResNets \cite{he2016deep} succeed in building very deep CNNs
with the help of residual connections.
ResNeXts \cite{xie2017aggregated} and Res2Nets \cite{gao2021res2net}
improve ResNets \cite{he2016deep} by exploring its cardinal operation.
DenseNets \cite{huang2017densely} introduce dense connections that
connect each layer to all its subsequent layers for easing optimization.
MobileNets \cite{howard2017mobilenets,sandler2018mobilenetv2}
decompose a vanilla convolution into a $1\times 1$ convolution
and a depthwise separable convolution to build lightweight CNNs
for mobile and embedded vision applications.
ShuffleNets \cite{ma2018shufflenet,zhang2018shufflenet} further reduce 
the latency of MobileNets \cite{howard2017mobilenets,sandler2018mobilenetv2}
by replacing the $1\times 1$ convolution with the grouped $1\times 1$ 
convolution and the channel shuffle operation.
EfficientNet \cite{tan2019efficientnet} and MnasNet \cite{tan2019mnasnet}
adopt neural architecture search (NAS) to search for optimal
network architectures.
Since our work focuses on the transformer \cite{vaswani2017attention},
a comprehensive survey of CNNs is beyond the scope of this paper.
Please refer to \cite{khan2020survey} and \cite{liu2017survey} for 
a more extensive survey.

\subsection{Vision Transformer}
Transformer is initially proposed for machine translation 
in NLP \cite{vaswani2017attention}.
Through MHSA, transformer entirely relies on self-attention 
to model global token-to-token dependencies.
Considering that global relationship is also highly required by computer 
vision tasks, it is a natural idea to adopt transformer 
for improving vision tasks.
However, transformer is designed to process sequence data and
thus cannot process images directly.
Hence, some researchers use CNNs to extract 2D representation that 
is then flattened and fed into the transformer
\cite{carion2020end,zhu2020deformable,wang2021max,liu2021end,hu2021istr}.
DETR \cite{carion2020end} is a milestone in this direction.

Instead of relying on the CNN backbone for feature extraction,
Dosovitskiy \etal \cite{dosovitskiy2021image} proposed the first
vision transformer (ViT).
They split an image into small patches, and each patch is viewed as
a word/token in an NLP application.
Thus, a pure transformer network can be directly adopted  
with the class token used for image classification.
Their method achieved competitive performance on ImageNet 
\cite{russakovsky2015imagenet}.
Then, DeiT \cite{touvron2021training} alleviates the required resources 
for training ViT \cite{dosovitskiy2021image} via knowledge distillation.
T2T-ViT \cite{yuan2021tokens} proposes to split an image with 
overlapping to better preserve local structures.
CvT \cite{wu2021cvt} introduces depthwise convolution for generating the query, key, and value in the computation of MHSA.
CPVT \cite{chu2021conditional} proposes to replace the absolute positional encoding with the conditional positional encoding via a depthwise convolution.
Some efforts are contributed to building the pyramid structure for 
the vision transformer using pooling operations
\cite{heo2021rethinking,liu2021swin,wang2021pyramid,fan2021multiscale}.
Among them, PVT \cite{wang2021pyramid} and MViT \cite{fan2021multiscale} first adopt the single pooling operation
to reduce the number of tokens when computing MHSA.
In this way, they actually conduct token-to-region relationship modeling,
not the expected token-to-token modeling.
Since this paper also resolves the long sequence length problem through pooling, we adopt PVT \cite{wang2021pyramid} and MViT \cite{fan2021multiscale}
as strong baselines in this paper.
Swin Transformer \cite{liu2021swin} reduces the computational load 
of MHSA by computing it within small windows.
However, Swin Transformer gradually achieves global relationship 
modeling by window shift, somewhat like CNNs that 
enlarge the receptive field by stacking more layers \cite{han2021demystifying}.
Hence, we think that Swin Transformer \cite{liu2021swin} sacrifices 
an essential characteristic of the vision transformer, 
\ie, \textit{direct global relationship modeling}.

Different from PVT \cite{wang2021pyramid} and MViT \cite{fan2021multiscale}
where the pooled feature extracted by a single pooling operation seems less powerful,
we adapt the idea of pyramid pooling to the vision transformer, 
simultaneously reducing the sequence length and learning powerful contextual representations.
With more powerful representations, it is intuitive that pyramid pooling may work better than single pooling for computing self-attention in MHSA.
Pyramid pooling is very efficient and thus will only induce negligible computational cost.
Experiments show that the proposed P2T performs substantially better performance than previous CNN- and transformer-based networks.
Besides, our design is also compatible with other transformer techniques
such as patch embedding \cite{jiang2021token}, 
positional encoding \cite{chu2021conditional}, 
and feed-forward network \cite{li2021localvit,yuan2021incorporating,liu2021transformer}.

\subsection{Pyramid Pooling}
In computer vision, pyramid pooling is a long-history and widely-acknowledged technique for extracting feature presentations. 
Before the renaissance of deep CNNs \cite{krizhevsky2012imagenet}, there emerged several well-known works that applied pyramid pooling for recognizing natural scenes \cite{grauman2005pyramid,lazebnik2006beyond}. 
Inspired by \cite{grauman2005pyramid,lazebnik2006beyond},
He \etal \cite{he2015spatial} introduced pyramid pooling to deep CNNs for image classification and object detection.
They adopted several pooling operations to pool the final convolutional feature map of a CNN backbone into several fixed-size maps.
These resulting maps are then flattened and concatenated into a fixed-length representation for robust visual recognition.
Then, Zhao \etal \cite{zhao2017pyramid} applied pyramid pooling for semantic segmentation.
Instead of flattening in \cite{he2015spatial}, they upsampled the pooled fixed-size maps into the original size and concatenated the upsampled maps for prediction.
Their success suggests the effectiveness of pyramid pooling in dense prediction.
After that, pyramid pooling has been widely applied to various vision tasks such as semantic segmentation \cite{zhao2017pyramid,chen2017deeplab,sarker2018slsdeep,yuan2021ocnet,lian2021cascaded} and object detection \cite{he2015spatial,yoo2015multi,kim2018parallel,huang2020dc}.

Unlike existing literature that explores pyramid pooling in CNNs for specific tasks,
we propose to adapt the concept of pyramid pooling
to the vision transformer backbone network.
With this idea, we first embed the pyramid pooling into the basic pooling-based attention block of our P2T backbone, which can simultaneously reduce the sequence length and learn powerful contextual feature representations.
P2T can be easily used by various vision tasks for feature representation learning, while previous works about pyramid pooling are limited to a specific vision task.
Extensive experiments on image classification, semantic segmentation, object detection, and instance segmentation demonstrate the superiority of P2T compared with existing CNN- and transformer-based networks.
Therefore, this work is distinctive and would benefit the research on various vision tasks.

\CheckRmv{
\begin{figure}[!t]
    \centering
    \includegraphics[width=0.95\columnwidth]{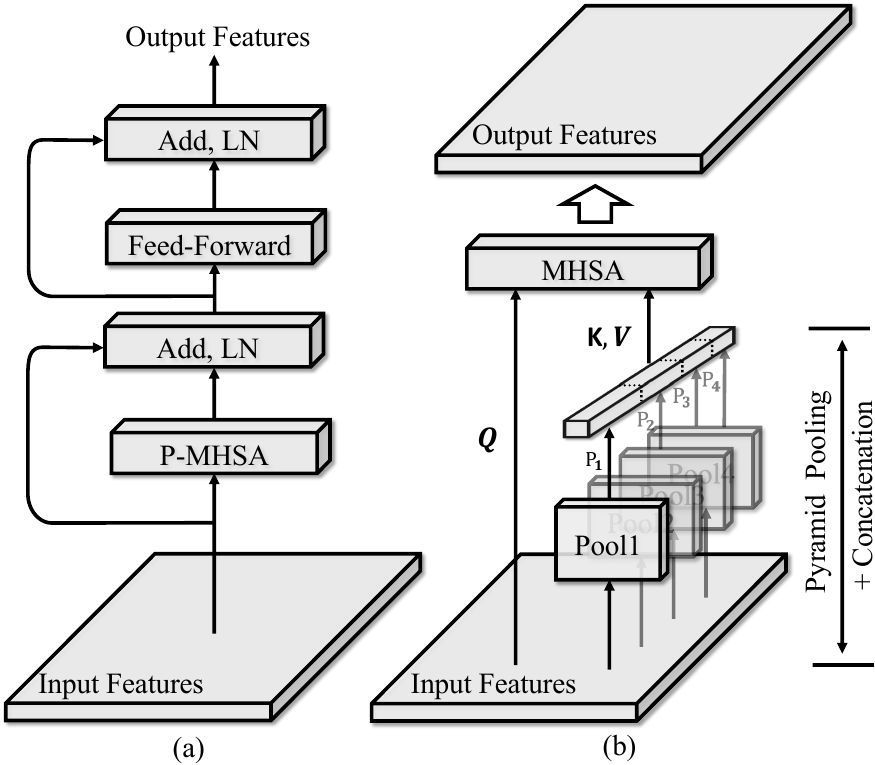}
    \caption{\textbf{Illustration of Pyramid Pooling Transformer.} 
    (a) The brief structure of Pyramid Pooling Transformer.
    (b) The detailed structure of the pooling-based MHSA.}
    \label{fig:p2t-arch}
\end{figure}
}

\section{Methodology} \label{sec:method}
In this section, we first provide an overview of our P2T networks
in \secref{sec:network}. 
Then, we present the architecture of P2T with pooling-based MHSA 
in \secref{sec:p2t}.
Finally, we introduce some implementation details of our networks
in \secref{sec:details}.

\subsection{Overview} \label{sec:network}
The overall architecture of P2T is illustrated in~\figref{fig:pipeline}.
With a natural color image as input, P2T first splits it into $\frac{H}{4}\times\frac{W}{4}$ patches, each flattened to 
48 ($4\times 4\times 3$) elements.
Following \cite{wang2021pyramid}, we feed these flattened patches to 
a patch embedding module, which consists of a linear projection layer 
followed by the addition with a learnable positional encoding.
The patch embedding module will expand the feature dimension 
from 48 to $C_1$.
Then, we stack the proposed pyramid pooling transformer blocks
that will be introduced in \secref{sec:p2t}.
The whole network can be divided into four stages with feature dimensions
of $C_i$ $(i=\{1,2,3,4\})$, respectively.
Between every two stages, each $2\times 2$ patch group is concatenated
and linearly projected from $4\times C_i$ to $C_{i+1}$ dimension 
($i=\{1,2,3\}$).
In this way, the scales of four stages become 
$\frac{H}{4}\times\frac{W}{4}$, $\frac{H}{8}\times\frac{W}{8}$,
$\frac{H}{16}\times\frac{W}{16}$, 
and $\frac{H}{32}\times\frac{W}{32}$, respectively.
From four stages, we can derive four feature representations
$\{\mathbf{B}_1, \mathbf{B}_2, \mathbf{B}_3, \mathbf{B}_4\}$, 
respectively.
Only $\mathbf{B}_4$ will be used for final prediction for image classification, 
while all pyramid features can be utilized for downstream scene understanding tasks.

\newcommand{\splitcell}[1]{\begin{tabular}{@{}c@{}}#1\end{tabular}}
\newcommand{\bsplitcell}[1]{$\left[\splitcell{#1}\right]$}

\begin{table*}[!tb]
    \centering
    \renewcommand{\arraystretch}{1.2}
    \renewcommand{\tabcolsep}{4.mm}
    \caption{\textbf{Detailed settings of the proposed P2T.}
    The parameters of building blocks are shown in brackets, 
    with the numbers of blocks stacked.
    For the first stage, we apply a $7\times 7$ convolution with $C$ output channels and a stride of $S$ for patch embedding.
    Each IRB uses an expansion ratio of $E$.
    For simplicity, we omit the patch embedding operation, \ie, a $3\times 3$ convolution with a stride of $S=2$, after the $t$-th stage 
    ($t=\{2,3,4\}$).
    ``\#Params'' refers to the number of parameters. 
    ``\#Flops'' denotes the computational cost with the input size of $224\times 224$.}
    \label{tab:hyper_param_p2t}
    \begin{tabular}{c|c|c|c|c|c|c} \Xhline{0.4mm}
        Stage & Input Size & Operator & P2T-Tiny & P2T-Small & P2T-Base & P2T-Large
        \\ \hline
        1 & $224\times 224$ & $7\times 7$ conv.
        & $C=48$, $S=4$
        & \multicolumn{3}{c}{$C=64$, $S=4$}
        \\ \hline
        2 & $56\times 56$ & \splitcell{P-MHSA \\ IRB}
        & \bsplitcell{$C=48$\\ $E=8$} $\times$ 2
        & \bsplitcell{$C=64$\\ $E=8$} $\times$ 2
        & \bsplitcell{$C=64$\\ $E=8$} $\times$ 3
        & \bsplitcell{$C=64$\\ $E=8$} $\times$ 3
        \\ \hline
        3 & $28\times 28$ & \splitcell{P-MHSA \\ IRB}
        & \bsplitcell{$C=96$\\ $E=8$} $\times$ 2
        & \bsplitcell{$C=128$\\ $E=8$} $\times$ 2
        & \bsplitcell{$C=128$\\ $E=8$} $\times$ 4
        & \bsplitcell{$C=128$\\ $E=8$} $\times$ 8
        \\ \hline
        4 & $14\times 14$ & \splitcell{P-MHSA \\ IRB}
        & \bsplitcell{$C=240$\\ $E=4$} $\times$ 6
        & \bsplitcell{$C=320$\\ $E=4$} $\times$ 9
        & \bsplitcell{$C=320$\\ $E=4$} $\times$ 18
        & \bsplitcell{$C=320$\\ $E=4$} $\times$ 27
        \\ \hline
        5 & $7\times 7$ & \splitcell{P-MHSA \\ IRB}
        & \bsplitcell{$C=384$\\ $E=4$} $\times$ 3
        & \bsplitcell{$C=512$\\ $E=4$} $\times$ 3
        & \bsplitcell{$C=512$\\ $E=4$} $\times$ 3
        & \bsplitcell{$C=640$\\ $E=4$} $\times$ 3
        \\ \hline
        & $1\times 1$ & - 
        & \multicolumn{4}{c}{Global Average Pooling, 1000-d FC, Softmax}
        \\ \hline
        \multicolumn{3}{c|}{\#Params} & 11.6M & 24.1M & 36.1M & 54.5M
        \\ \hline
        \multicolumn{3}{c|}{\#Flops} & 1.8G & 3.7G & 6.5G & 9.8G
        \\ 
        \Xhline{0.4mm}
    \end{tabular}
\end{table*}

\subsection{Pyramid Pooling Transformer}\label{sec:p2t}
Pyramid pooling has been widely used in many scene 
understanding tasks collaborating with CNNs
\cite{he2015spatial,zhao2017pyramid,zhang2017amulet,chang2018pyramid,Liu21PamiPoolNet, wu2021regularized,park2018efficient,zhang2019image,zhang2018densely,wang2017stagewise}.
However, existing literature usually applies pyramid pooling on top of
CNN backbones for extracting global and contextual information
for a specific task.
In contrast, this paper is the first to explore pyramid pooling 
in transformers and backbone networks, targeting for improving
various scene understanding tasks generically.
To this end, we adapt the idea of pyramid pooling to the transformer,
simultaneously reducing the computational load of MHSA and
capturing rich contextual information.

Let us continue by introducing the proposed P2T, the structure of which
is illustrated in \figref{fig:p2t-arch} (a).
The input first passes through the pooling-based MHSA, 
whose output is added with the residual identity, followed by 
LayerNorm \cite{ba2016layer}.
Like the traditional transformer block
\cite{dosovitskiy2021image,touvron2021training,wang2021pyramid},
a feed-forward network (FFN)
follows for feature projection.
A residual connection and LayerNorm \cite{ba2016layer} are applied again.
The above process can be formulated as 
\begin{eqnarray}\label{eq:global_mix}
\begin{aligned}
  \mathbf{X}_{att} &= \text{LayerNorm}(\mathbf{X} + \text{P-MHSA}(\mathbf{X})),\\
  \mathbf{X}_{out} &= \text{LayerNorm}(\mathbf{X}_{att} + \text{FFN}(\mathbf{X}_{att})),
\end{aligned}
\end{eqnarray}
where $\mathbf{X}$, $\mathbf{X}_{att}$, and $\mathbf{X}_{out}$ 
are the input, the output of MHSA, and the output of 
the transformer block, respectively.
$\textbf{P-MHSA}$ is the abbreviation of pooling-based MHSA.

\subsubsection{Pooling-based MHSA}
Here, we present the design of our pooling-based MHSA.
Its structure is shown in \figref{fig:p2t-arch} (b).
First, the input $\mathbf{X}$ is reshaped into the 2D space.
Then, we apply multiple average pooling layers with various ratios onto 
the reshaped $\mathbf{X}$ to generate pyramid feature maps, like
\begin{eqnarray}\label{eq:pool}
\begin{aligned}
  \mathbf{P}_1 &= \text{AvgPool}_1(\mathbf{X}),\\
  \mathbf{P}_2 &= \text{AvgPool}_2(\mathbf{X}),\\
  &\cdots,\\
  \mathbf{P}_n &= \text{AvgPool}_n(\mathbf{X}),
\end{aligned}
\end{eqnarray}
where $\{\mathbf{P}_1, \mathbf{P}_2, ..., \mathbf{P}_n\}$ 
denote the generated pyramid feature maps and $n$ is 
the number of pooling layers.
Next, we feed pyramid feature maps to the depthwise convolution for relative positional encoding:
\begin{eqnarray}\label{eq:position}
  \mathbf{P}_i^{enc} = \text{DWConv}(\mathbf{P}_i) + \mathbf{P}_i,
  \quad i = 1, 2, \cdots, n,
\end{eqnarray}
where $\text{DWConv}(\cdot)$ indicates the depthwise convolution with the kernel size $3\times 3$, and $\mathbf{P}_i^{enc}$ is $\mathbf{P}_i$ with the relative positional encoding.
Since $\mathbf{P}_i$ is the pooled feature, the operation in \equref{eq:position} only has a little computational cost.
After that, we flatten and concatenate these pyramid feature maps:
\begin{eqnarray}
  \mathbf{P} = \text{LayerNorm}(\text{Concat}(\mathbf{P}_1^{enc}, \mathbf{P}_2^{enc}, ..., \mathbf{P}_n^{enc})),
\end{eqnarray}
where the flattening operation is omitted for simplicity.
In this way, $\mathbf{P}$ can be a shorter sequence than 
the input $\mathbf{X}$ if pooling ratios are large enough.
Besides, $\mathbf{P}$ contains the contextual abstraction of 
the input $\mathbf{X}$ and can thus serve as a strong substitute for the input $\mathbf{X}$
when computing MHSA.

Suppose the query, key, and value tensors in MHSA \cite{dosovitskiy2021image}
are $\mathbf{Q}$, $\mathbf{K}$, and $\mathbf{V}$, respectively.
Instead of using traditional 
\begin{eqnarray}
(\mathbf{Q},\mathbf{K},\mathbf{V})=(\mathbf{X}\mathbf{W}^q,\mathbf{X}\mathbf{W}^k,\mathbf{X}\mathbf{W}^v),
\end{eqnarray}
we propose to use 
\begin{eqnarray}
(\mathbf{Q},\mathbf{K},\mathbf{V})=(\mathbf{X}\mathbf{W}^q,\mathbf{P}\mathbf{W}^k,\mathbf{P}\mathbf{W}^v),
\end{eqnarray}
in which $\mathbf{W}^q$, $\mathbf{W}^k$, and $\mathbf{W}^v$ denote
the weight matrices of linear transformations for generating 
the query, key, and value tensors, respectively.
Then, $\mathbf{Q},\mathbf{K},\mathbf{V}$ are fed into the
attention module to compute the attention $\mathbf{A}$, 
which can be formulated as below:
\begin{equation}\label{eq:attention}
\mathbf{A} = \text{Softmax}(\frac{\mathbf{Q}\times \mathbf{K}^\mathbf{T}}{\sqrt{d_K}})\times \mathbf{V},
\end{equation}
where $d_K$ is the channel dimension of $\mathbf{K}$,
and $\sqrt{d_K}$ can serve as an approximate normalization.
The $\text{Softmax}$ function is applied along the rows of the matrix.
\equref{eq:attention} omits the concept of multiple heads 
\cite{vaswani2017attention,dosovitskiy2021image}
for simplicity.

Since $\mathbf{K}$ and $\mathbf{V}$ have a smaller length
than $\mathbf{X}$, the proposed P-MHSA is more efficient 
than traditional MHSA \cite{vaswani2017attention,dosovitskiy2021image}.
Besides, since $\mathbf{K}$ and $\mathbf{V}$ contains
highly-abstracted multi-scale information, the proposed P-MHSA
has a stronger capability in global contextual dependency modeling,
which is helpful for scene understanding 
\cite{zhao2017pyramid,he2015spatial,park2018efficient,chang2018pyramid,zhang2019image,zhang2018densely,wang2017stagewise}.
From a different perspective, pyramid pooling is usually used 
as an effective technique connected upon backbone networks;
in contrast, this paper first exploits pyramid pooling within
backbone networks through transformers, thus providing powerful 
feature representation learning for scene understanding.
With the above analyses, P-MHSA is expected to be more efficient
and more effective than traditional MHSA
\cite{vaswani2017attention,dosovitskiy2021image}.

\myPara{Analysis of computational complexity.}
As described in \equref{eq:pool},
the proposed pooling-based attention leverages several pooling operations to generate pyramid feature maps.
The pyramid pooling operation only has negligible $O(NC)$ computational complexity, where $N$ and $C$ represent the sequence length and the feature dimension, respectively.
Hence, the computational complexity for computing self-attention can be formulated as
\begin{equation}
    O(\text{P-MHSA}) = (N+2M)C^2 + 2NMC,
\end{equation}
where $M$ is the concatenated sequence length of all pooled features.
For the default pooling ratios of \{12, 16, 20, 24\}, we have $M\approx\frac{N}{66.3}\approx \frac{N}{8^2}$, which is comparable with the computational cost of MHSA in PVT \cite{wang2021pyramid}.

\subsubsection{Feed-Forward Network}\label{sec:ffn}
Feed-Forward Network (FFN) is an essential component of transformers
for feature enhancement \cite{vaswani2017attention,dong2021attention}.
Previous transformers usually apply an MLP as the FFN
\cite{vaswani2017attention,dosovitskiy2021image,wang2021pyramid}
and entirely rely on attention to capture inter-pixel dependencies.
Though effective, this architecture is not good at learning 
2D locality, %
which plays a critical role in scene understanding.
To this end, we follow \cite{li2021localvit,yuan2021incorporating} to insert the depthwise convolution into FFN so that the resulting transformer can 
inherit the merits of both transformer (\ie, long-range dependency
modeling) and CNN (\ie, 2D locality).
Specifically, we adopt the Inverted Bottleneck Block (IRB),
proposed in MobileNetV2 \cite{sandler2018mobilenetv2},
as the FFN.

To adapt IRB for the vision transformer, we first transform the input 
sequence $\mathbf{X}_{att}$ to a 2D feature map $\mathbf{X}_{att}^I$:
\begin{equation}
    \mathbf{X}_{att}^I = \text{Seq2Image}(\mathbf{X}_{att}),
\end{equation}
where $\text{Seq2Image}(\cdot)$ is to reshape the 1D sequence 
to a 2D feature map.
Given the input $\mathbf{X}_{att}^I$, IRB can be directly applied, like
\begin{eqnarray}\label{eq:irb}
\begin{aligned}
  \mathbf{X}_{\text{IRB}}^{1} &= \text{Act}(\mathbf{X}_{att}^I \mathbf{W}_{\text{IRB}}^1), \\
  \mathbf{X}_{\text{IRB}}^{\text{out}} &= \text{Act}(\text{DWConv}(\mathbf{X}_{\text{IRB}}^{1}))\mathbf{W}_{\text{IRB}}^2, 
\end{aligned}
\end{eqnarray}
where $\mathbf{W}_{\text{IRB}}^1$, $\mathbf{W}_{\text{IRB}}^2$ 
indicate the weight matrices of $1\times 1$ convolutions,
``Act'' indicates the nonlinear activation function,
$\mathbf{X}_{\text{IRB}}^{\text{out}}$ is the output of IRB.
Since $\mathbf{X}_{\text{IRB}}^{\text{out}}$ is a 2D feature map, 
we finally transform it to a 1D sequence:
\begin{equation}
    \mathbf{X}_{\text{IRB}}^S = \text{Image2Seq}(\mathbf{X}_{\text{IRB}}^{\text{out}}),
\end{equation}
where $\text{Image2Seq}(\cdot)$ is the operation that reshapes 
the 2D feature map to a 1D sequence.
$\mathbf{X}_{\text{IRB}}^S$ is the output of FFN,
with the same shape as $\mathbf{X}_{att}$.

\subsection{Implementation Details}\label{sec:details}
\parahead{P2T with different depths.}
Following previous backbone architectures 
\cite{he2016deep,xie2017aggregated,wang2021pyramid,liu2021swin,liu2021transformer}, 
we build P2T with different depths via stacking the different number 
of pyramid pooling transformers at each stage.
In this manner, we propose four versions of P2T, \ie,
P2T-Tiny, P2T-Small, P2T-Base, and P2T-Large with similar numbers of
parameters to ResNet-18 \cite{he2016deep}, ResNet-50 \cite{he2016deep}, 
ResNet-101 \cite{he2016deep}, and PVT-Large \cite{wang2021pyramid}, respectively.
Each head of P-MHSA has 64 feature channels except that each head has 48 feature channels in P2T-Tiny.
Other configurations for different versions of P2T 
are shown in \tabref{tab:hyper_param_p2t}.

\myPara{Pyramid pooling settings.}
We empirically set the number of parallel pooling operations 
in P-MHSA as 4.
At different stages, the pooling ratios of pyramid pooling 
transformers are different.
The pooling ratios for the first stage are empirically set as 
$\{12, 16, 20, 24\}$.
Pooling ratios in each next stage are divided by 2 
except that, in the last stage, they are set as $\{1, 2, 3, 4\}$.
In each transformer block, all depthwise convolutions (\equref{eq:position}) in P-MHSA share the same parameters.

\myPara{Other settings.}
Although a larger kernel size (\eg, $5\times 5$) of depthwise 
convolution (in \equref{eq:position}) can bring better performance, the kernel size of all 
depthwise convolutions is set to $3\times 3$ for efficiency.
We choose Hardswish \cite{howard2019searching} as the nonlinear 
activation function because it saves much memory compared with 
GELU \cite{hendrycks2016gaussian}.
Hardswish \cite{howard2019searching} also empirically works well.
Same with PVTv2 \cite{wang2021pvtv2}, we apply overlapped patch embedding. That is, we use $3\times 3$ convolution with a stride of 2 for patch embedding from the second to the last stage, while we apply a $7\times 7$ convolution with a stride of 4 for patch embedding in the first stage.

\section{Experiments}
We first introduce the experiments on image classification in \secref{sec:exp_classification}. 
Then we validate P2T's effectiveness on several scene understanding tasks, \ie, semantic segmentation, 
object detection, and instance segmentation 
in \secref{sec:exp_sematicseg}, \secref{sec:exp_detection}, and \secref{sec:exp_instance}, respectively.
At last, we conduct ablation studies for better understanding 
our method in \secref{sec:ablation}.

\begin{table}[!t]
    \centering
    \setlength{\tabcolsep}{1mm}
    \caption{\textbf{Image classification results on the ImageNet-1K dataset
    \cite{russakovsky2015imagenet}.}
    ``Top-1'' indicates the top-1 accuracy rate. 
    ``*'' indicates the results with knowledge distillation \cite{touvron2021training}.
    ``\#P (M)'' denotes the number of parameters (M).
    Both the number of computational cost (GFlops) and running speed (frames per second, FPS) are reported with the default $224\times 224$ input size for each network except that the speed of ViT-B \cite{dosovitskiy2021image} is tested with the input size of $384 \times 384$. FPS is tested on a single RTX 2070 GPU. The results of the proposed P2T are marked in \textbf{bold}.}
    \label{tab:exp_cls}
    \resizebox{\columnwidth}{!}{%
    \begin{tabular}{l|c|c|c|c} \Xhline{1pt}
        Method & \#P (M) $\downarrow$ & GFlops $\downarrow$ & Top-1 (\%) $\uparrow$ & FPS $\uparrow$ \\ \Xhline{1pt}
        ResNet-18~\cite{he2016deep} & 11.7 & 1.8 & 68.5 & 1410 \\
        DeiT-Tiny/16*~\cite{touvron2021training} & 5.7 & 1.3 & 72.2 & 1212 \\
        ViL-Tiny~\cite{zhang2021multi} & 6.7 & 1.3 & 76.7 & 441 \\
        PVT-Tiny~\cite{wang2021pyramid} & 13.2 & 1.9 & 75.1 & 608 \\
        PVTv2-B1~\cite{wang2021pvtv2} & 13.1 & 2.1 & 78.7 & 502  \\
        \rowcolor{mygray} \textbf{P2T-Tiny (Ours)} & \textbf{11.6} 
        & \textbf{1.8} & \textbf{79.8} & \textbf{473} \\ \hline
        ResNet-50~\cite{he2016deep} & 25.6 & 4.1 & 78.5 & 483 \\
        ResNeXt-50-32x4d~\cite{xie2017aggregated} & 25.0 & 4.3 & 79.5 & 407 \\
        Res2Net-50~\cite{gao2021res2net} & 25.7 & 4.5 & 80.3 & 430 \\
        DeiT-Small/16*~\cite{touvron2021training} & 22.1 & 4.6 & 79.9 & 489 \\
        PVT-Small~\cite{wang2021pyramid} & 24.5 & 3.8 & 79.8 & 336 \\
        T2T-ViT$_t$-14 \cite{yuan2021tokens} & 21.5 & 5.2 & 80.7 & 305 \\
        Swin-T~\cite{liu2021swin} & 29.0 & 4.5 & 81.3 & 349 \\
        Twins-SVT-S \cite{chu2021twins} & 24.0 & 2.9 & 81.7 & 439 \\
        ViL-Small~\cite{zhang2021multi} & 25.0 & 4.9 & 82.4 & 187 \\
        PVTv2-B2~\cite{wang2021pvtv2} & 25.4 & 4.0 & 82.0 & 284 \\
        \rowcolor{mygray} \textbf{P2T-Small (Ours)} & \textbf{24.1} 
        & \textbf{3.7} & \textbf{82.4} & \textbf{284} \\ \hline
        ResNet-101~\cite{he2016deep} & 44.7 & 7.9 & 79.8 & 288 \\
        ResNeXt-101-32x4d~\cite{xie2017aggregated} & 44.2 & 8.0 & 80.6 & 228 \\
        Res2Net-101~\cite{gao2021res2net} & 45.2 & 8.3 & 81.2 & 265 \\
        PVT-Medium~\cite{wang2021pyramid} & 44.2 & 6.7 & 81.2 & 216 \\
        T2T-ViT$_t$-19 \cite{yuan2021tokens} & 39.2 & 8.4 & 81.4 & 202 \\
        Swin-S~\cite{liu2021swin} & 50.0 & 8.7 & 83.0 & 207 \\
        ViL-Medium~\cite{zhang2021multi} & 40.4 & 8.7 & 83.5 & 114 \\  
        MViT-B-16~\cite{fan2021multiscale} & 37.0 & 7.8 & 83.1 & 222 \\
        PVTv2-B3~\cite{wang2021pvtv2} & 45.2 & 6.9 & 83.2 & 189  \\
        \rowcolor{mygray} \textbf{P2T-Base (Ours)} & \textbf{36.1}
        & \textbf{6.5} & \textbf{83.5} & \textbf{182} \\ \Xhline{1pt}
        ResNeXt-101-64x4d~\cite{xie2017aggregated} & 83.5 & 15.6 & 81.5 & 147 \\
        MViT-B-24~\cite{fan2021multiscale} & 53.5 & 10.9 & 83.0 & 151 \\
        ViL-Base \cite{zhang2021multi} & 57.0 & 13.4 & 83.7 & 67 \\
        PVT-Large~\cite{wang2021pyramid} & 61.4 & 9.8 & 81.7 & 152 \\
        DeiT-Base/16*~\cite{touvron2021training} & 86.6 & 17.6 & 81.8 & 161 \\
        ViT-Base/16~\cite{dosovitskiy2021image} & 86.6 & 17.6 & 77.9 & 49 \\
        Swin-B~\cite{liu2021swin} & 88.0 & 15.4 & 83.3 & 140\\
        Twins-SVT-L~\cite{chu2021twins} & 99.2 & 14.8 & 83.3 & 143 \\
        PVTv2-B4~\cite{wang2021pvtv2} & 62.6 & 10.1 & 83.6 & 133\\
        PVTv2-B5~\cite{wang2021pvtv2} & 82.0 & 11.8 & 83.8 & 120 \\
        \rowcolor{mygray} \textbf{P2T-Large (Ours)} & \textbf{54.5}
        & \textbf{9.8} & \textbf{83.9} & \textbf{128} \\ \Xhline{1pt}
    \end{tabular}}
\end{table}

\begin{table}[!t]
    \centering
    \setlength{\tabcolsep}{1mm}
    \caption{\textbf{Experimental results on the validation set of the ADE20K 
    dataset \cite{zhou2017scene} for semantic segmentation.}
    We replace the backbone of Semantic FPN \cite{kirillov2019panoptic} with various network architectures.
    The number of GFlops is calculated with the input size of $512\times 512$.
    FPS is tested on a single RTX 2070 GPU.
    The results of P2T backbones are marked in \textbf{bold}.}
    \label{tab:exp_ade20k}
	\resizebox{\columnwidth}{!}{%
    \begin{tabular}{l|c|c|c|c} \Xhline{1pt}
        \multirow{2}[0]{*}{Backbone}
        & \multicolumn{4}{c}{Semantic FPN \cite{kirillov2019panoptic}}
        \\ \cline{2-5}
        & \#Param (M) $\downarrow$ & GFlops $\downarrow$ & mIoU (\%) $\uparrow$  & FPS $\uparrow$
        \\ \Xhline{1pt}
        ResNet-18 \cite{he2016deep} & 15.5 & 31.9 & 32.9 & 68 \\
        PVT-Tiny \cite{wang2021pyramid} & 17.0 & 32.1 & 35.7 & 36 \\
        PVTv2-B1~\cite{wang2021pvtv2} & 17.8 & 33.1 & 41.5 & 30 \\
        \rowcolor{mygray} \textbf{P2T-Tiny (Ours)} & \textbf{15.4} & \textbf{31.6}  & \textbf{43.4} & \textbf{31} \\ \hline
        ResNet-50 \cite{he2016deep} & 28.5 & 45.4  & 36.7 & 35 \\
        PVT-Small \cite{wang2021pyramid} & 28.2 & 42.9 & 39.8 & 26 \\
        Swin-T~\cite{liu2021swin} & 31.9 & 46 & 41.5 & 26 \\
        Twins-SVT-S~\cite{chu2021twins} & 28.3  & 37 & 43.2 & 27 \\
        PVTv2-B2~\cite{wang2021pvtv2} & 29.1 & 44.1 & 46.1 & 21 \\
        \rowcolor{mygray} \textbf{P2T-Small (Ours)} & \textbf{27.8}
        & \textbf{42.7} & \textbf{46.7} & \textbf{24} \\ \hline
        ResNet-101 \cite{he2016deep} & 47.5 & 64.8 & 38.8 & 26 \\
        ResNeXt-101-32x4d \cite{xie2017aggregated} & 47.1 & 64.6 & 39.7 & 20 \\
        PVT-Medium \cite{wang2021pyramid} & 48.0 & 59.4 & 41.6 & 19 \\
        Swin-S \cite{liu2021swin} & 53.2 & 70 & 45.2 & 18 \\
        Twins-SVT-B \cite{chu2021twins} & 60.4 & 67 & 45.3 & 17 \\
        PVTv2-B3 \cite{wang2021pvtv2} & 49.0 & 60.7 & 47.3 & 15 \\ 
        \rowcolor{mygray} \textbf{P2T-Base (Ours)} & \textbf{39.8}
        & \textbf{58.5} & \textbf{48.7} & \textbf{16} \\\Xhline{1pt}
        ResNeXt-101-64x4d \cite{xie2017aggregated} & 86.4 & 104.2 & 40.2 & 15 \\
        PVT-Large \cite{wang2021pyramid} & 65.1 & 78.0 & 42.1 & 15 \\
        Swin-B \cite{liu2021swin} & 91.2 & 107 & 46.0 & 13\\
        Twins-SVT-L \cite{chu2021twins} & 102 & 103.7 & 46.7 & 13\\
        PVTv2-B4 \cite{wang2021pvtv2} & 66.3 & 79.6 & 48.6 & 11 \\
        PVTv2-B5 \cite{wang2021pvtv2} & 85.7 & 89.4 & 48.9 & 10\\
        \rowcolor{mygray} \textbf{P2T-Large (Ours)} & \textbf{58.1}
        & \textbf{77.7} & \textbf{49.4} & \textbf{12} \\ 
        \Xhline{1pt}
    \end{tabular}}
\end{table}

\begin{table*}
\setlength{\tabcolsep}{0.4mm}
\caption{\textbf{Object detection results with RetinaNet \cite{lin2017focal} and instance segmentation results with Mask R-CNN \cite{he2020mask}
    on the MS-COCO \texttt{val2017} set \cite{lin2014microsoft}.}
    ``R'' and ``X'' represent the ResNet \cite{he2016deep} and 
    ResNeXt \cite{xie2017aggregated}, respectively.
    The number of Flops is computed with the input size of $800 \times 1280$. 
    FPS is tested on a single RTX 2070 GPU.
    The results of P2T backbones are marked in \textbf{bold}.}
\label{tab:det_seg}
\resizebox{\textwidth}{!}{%
\begin{tabular}{l||ccc|lcc|lcc||ccc|lcc|lcc}
\Xhline{1pt}
\multirow{3}{*}{Backbone} & \multicolumn{9}{c||}{Object Detection} & \multicolumn{9}{c}{Instance Segmentation} \\
\cline{2-19}
& \multirow{2}{*}{\tabincell{c}{\#Param\\(M) $\downarrow$}} & \multirow{2}{*}{\tabincell{c}{\#Flops\\(G)$\downarrow$}}   &
\multirow{2}{*}{\tabincell{c}{\#FPS\\$\uparrow$}}   &
\multicolumn{6}{c||}{RetinaNet \cite{lin2017focal}} & \multirow{2}{*}{\tabincell{c}{\#Param\\(M) $\downarrow$}} & \multirow{2}{*}{\tabincell{c}{\#Flops\\(G) $\downarrow$}} &
\multirow{2}{*}{\tabincell{c}{\#FPS\\$\uparrow$}}   &
\multicolumn{6}{c}{Mask R-CNN \cite{he2020mask}} \\
\cline{5-10} 
\cline{14-19}
& & &  &AP $\uparrow$ &AP$_{50}$ &AP$_{75}$ &AP$_S$ $\uparrow$ &AP$_M$ &AP$_L$ & &  &  &AP$^{\rm b}$ $\uparrow$ &AP$_{50}^{\rm b}$ &AP$_{75}^{\rm b}$  &AP$^{\rm m} $ $\uparrow$ &AP$_{50}^{\rm m}$ &AP$_{75}^{\rm m}$\\
\Xhline{1pt}
R-18~\cite{he2016deep} & 21.3 & 190 & 19.3 &  31.8 & 49.6 & 33.6 & 16.3 & 34.3 & 43.2  &31.2 & 209 & 17.3 & 34.0 & 54.0 & 36.7 & 31.2 & 51.0 & 32.7\\
ViL-Tiny~\cite{zhang2021multi} & 16.6 & 204 & 4.2 & 40.8 & 61.3 & 43.6 & 26.7 & 44.9 & 53.6 & 26.9 & 223 & 3.9 & 41.4 & 63.5 & 45.0 & 38.1 & 60.3 & 40.8 \\
PVT-Tiny~\cite{wang2021pyramid} & 23.0 & 205 & 10.7 & {36.7} & {56.9} & {38.9} & {22.6} & {38.8} & {50.0}  & 32.9 & 223 & 10.0 & {36.7} & {59.2} & {39.3} & {35.1} & {56.7} & {37.3} \\
PVTv2-B1~\cite{wang2021pvtv2} &23.8& 209 & 8.5 & 40.2& 60.7& 42.4& 22.8& 43.3& 54.0 & 33.7 & 227 & 8.0 & 41.8 & 64.3 & 45.9 & 38.8 & 61.2 & 41.6\\
\rowcolor{mygray}  \textbf{P2T-Tiny (Ours)} & \textbf{21.1} & \textbf{206} & \textbf{9.3} & \textbf{41.3} & \textbf{62.0}& \textbf{44.1} & \textbf{24.6} & \textbf{44.8} & \textbf{56.0} & \textbf{31.3} & \textbf{225} & \textbf{8.8} & \textbf{43.3} & \textbf{65.7} & \textbf{47.3} & \textbf{39.6} & \textbf{62.5} & \textbf{42.3}\\
\rowcolor{mygray}
\hline
R-50~\cite{he2016deep} & 37.7 & 239 & 13.0 & 36.3 & 55.3 & 38.6 & 19.3 & 40.0 & 48.8 & 44.2 & 260 & 11.5 & 38.0 & 58.6 & 41.4 & 34.4 & 55.1 & 36.7\\
PVT-Small~\cite{wang2021pyramid} & {34.2} & 261 & 7.7 & {40.4} & {61.3} & {43.0} & {25.0} & {42.9} & {55.7} &{44.1} & 280 & 7.0 &{40.4} & {62.9} & {43.8} & {37.8} & {60.1} & {40.3}\\
Swin-T \cite{liu2021swin} & 38.5 & 248 & 9.7 & 41.5& 62.1& 44.2& 25.1& 44.9& 55.5 & 47.8 & 264 & 8.8 & 42.2& 64.6& 46.2& 39.1& 61.6 &42.0\\
ViL-Small~\cite{zhang2021multi} & 35.7 & 292 & 3.4 & 44.2 & 65.2 & 47.6 & 28.8 & 48.0 & 57.8 & 45.0 & 310 & 3.2 & 44.9 & 67.1 & 49.3 & 41.0 & 64,2 & 44.1 \\
Twins-SVT-S~\cite{chu2021twins} &  34.3 & 236 & 8.5 & 43.0 & 64.2&46.3& 28.0& 46.4& 57.5 & 44.0 & 254 & 7.7 & 43.4 & 66.0& 47.3 & 40.3 & 63.2 & 43.4 \\
PVTv2-B2~\cite{wang2021pvtv2} & 35.1 & 266 & 5.8 & 43.8 &64.8 & 46.8 &26.0 &47.6 &59.2 & 45.0 & 285 & 5.4 & 45.3 & 67.1 & 49.6 & 41.2 & 64.2 & 44.4 \\
\rowcolor{mygray} \textbf{P2T-Small (Ours)} & \textbf{33.8} & \textbf{260} & \textbf{7.4} & \textbf{44.4} & \textbf{65.3} & \textbf{47.6} & \textbf{27.0} & \textbf{48.3} & \textbf{59.4} & \textbf{43.7} & \textbf{279} & \textbf{6.7} & \textbf{45.5} & \textbf{67.7} & \textbf{49.8} & \textbf{41.4} & \textbf{64.6} & \textbf{44.5} \\
\hline
R-101~\cite{he2016deep} &56.7 & 315 & 9.8 & 38.5 & 57.8 & 41.2 & 21.4 & 42.6 & 51.1  &63.2 & 336 & 9.1 & 40.4 & 61.1 & 44.2 & 36.4 & 57.7 & 38.8 \\
X-101-32x4d~\cite{xie2017aggregated} &56.4 & 319 & 8.5 & 39.9 & 59.6 & 42.7 & 22.3 & 44.2 & 52.5 & {62.8} & 340 & 7.9 & 41.9 & 62.5 & 45.9 & 37.5 & 59.4 & 40.2 \\
PVT-Medium~\cite{wang2021pyramid} & 53.9 & 349 & 5.7  & {41.9} & {63.1} & {44.3} & {25.0} & {44.9} & {57.6} &63.9 & 367 & 5.3  & {42.0} &{64.4} &45.6 &{39.0}& {61.6}& {42.1}\\
Swin-S~\cite{liu2021swin} & 59.8 & 336 & 7.1 & 44.5 & 65.7 & 47.5 & 27.4 & 48.0 & 59.9 &  69.1 & 354 & 6.6  & 44.8  & 66.6 & 48.9 & 40.9 & 63.4 & 44.2 \\
PVTv2-B3~\cite{wang2021pvtv2} &55.0 & 354 & 4.5 & 45.9 & 66.8 & 49.3 & 28.6 & 49.8 & 61.4 & 64.9 & 372 & 4.2 & 47.0 & 68.1 & 51.7 & 42.5 & 65.7 & 45.7 \\
\rowcolor{mygray} \textbf{P2T-Base (Ours)} & \textbf{45.8} & \textbf{344} & \textbf{5.0} & \textbf{46.1} & \textbf{67.5} & \textbf{49.6} & \textbf{30.2} & \textbf{50.6} & \textbf{60.9} & \textbf{55.7} & \textbf{363} & \textbf{4.7} & \textbf{47.2} & \textbf{69.3} & \textbf{51.6} & \textbf{42.7} & \textbf{66.1} & \textbf{45.9} \\
\hline
X-101-64x4d~\cite{xie2017aggregated} & 95.5 & 473 & 6.2 & 41.0 & 60.9 & 44.0 & 23.9 & 45.2 & 54.0 &101.9 & 493 & 5.7 & 42.8 & 63.8 & {47.3} & 38.4 & 60.6 & 41.3 \\
PVT-Large~\cite{wang2021pyramid} & 71.1 & 450 & 4.4  & {42.6} & {63.7} & {45.4} & {25.8} & {46.0} & {58.4} & 81.0 & 469 & 4.1 & {42.9}& {65.0} & 46.6 &{39.5}& {61.9}& {42.5}  \\
Twins-SVT-B~\cite{chu2021twins} & 67.0 & 376 & 5.1 & 45.3 & 66.7& 48.1 & 28.5 & 48.9 & 60.6 & 76.3 & 395 & 4.6 & 45.2 & 67.6 & 49.3 & 41.5 & 64.5 & 44.8 \\
PVTv2-B4~\cite{wang2021pvtv2} &72.3 & 457 & 3.4 & 46.1 & 66.9 & 49.2 & 28.4 & 50.0 & 62.2 & 82.2 & 475 & 3.2 & 47.5 & 68.7 & 52.0 & 42.7 & 66.1 & 46.1 \\
PVTv2-B5~\cite{wang2021pvtv2} & {91.7} & 514 & 3.2  & {46.2} & {67.1} & {49.5} & {28.5} & {50.0} & 62.5 & {101.6} & 532 & 3.0 & {47.4} & {68.6} & {51.9} & {42.5} & {65.7}& {46.0}\\
\rowcolor{mygray} \textbf{P2T-Large (Ours)} & \textbf{64.4} & \textbf{449} & \textbf{3.8} & \textbf{47.2} & \textbf{68.4} & \textbf{50.9} & \textbf{32.4} & \textbf{51.6} & \textbf{62.2} & \textbf{74.0} & \textbf{467} & \textbf{3.5} & \textbf{48.3} & \textbf{70.2} & \textbf{53.3}  & \textbf{43.5} & \textbf{67.3}  & \textbf{46.9} \\

\Xhline{1pt}
\end{tabular}
}
\end{table*}

\subsection{Image Classification}\label{sec:exp_classification}
Image classification is the most common task for evaluating the
capability of backbone networks.
It aims to assign a class label to each natural image input.
Many other tasks build on top of image classification via applying classification networks as the backbones for feature extraction.

\myPara{Experimental setup.}
As described in \secref{sec:network}, only the output feature 
$\mathbf{B}_4$ of the last stage is utilized here.
Following regular CNN networks
\cite{he2016deep,huang2017densely,gao2021res2net},
we append a global average pooling layer and a fully-connected layer
on top of $\mathbf{B}_4$ to obtain the final classification scores.
We train our network on the ImageNet-1K dataset \cite{russakovsky2015imagenet},
which has 1.28M training images and 50k validation images.
For a fair comparison, we follow PVT \cite{wang2021pyramid}
to adopt the same training protocols as DeiT \cite{touvron2021training} (without knowledge distillation),
which is a standard choice for training vision transformers.
Specifically, we use AdamW \cite{loshchilov2017decoupled} as the 
optimizer, with the initial learning rate of 1e-3, weight decay of 0.05, and a mini-batch of 
1024 images.
We train P2T for 300 epochs with the cosine learning rate decay strategy.
Images are resized to the size of $224\times 224$ for training and testing.
Models are warmed up for the first five epochs.
The data augmentation is also the same as 
\cite{wang2021pyramid,touvron2021training}.

\myPara{Experimental results.}
The quantitative comparisons are summarized in \tabref{tab:exp_cls}.
All models are trained and evaluated with the input size of $224\times 224$ except that we follow the official ViT \cite{dosovitskiy2021image} to train and evaluate it with the input size of $384 \times 384$.
P2T largely outperforms regular CNN models like ResNets\cite{he2016deep} and ResNeXts \cite{xie2017aggregated}.
For example, 
although the running time of P2T-Tiny/Small/Base/Large is 2.98/1.70/1.58/1.15 times that of ResNet-18/50/101 \cite{he2016deep} and ResNeXt-101-64x4d \cite{xie2017aggregated},
the top-1 accuracy of P2T-Tiny/Small/Base/Large is 11.3\%/3.9\%/3.7\%/2.4\% better than that of ResNet-18/50/101 \cite{he2016deep} and ResNeXt-101-64x4d \cite{xie2017aggregated}, respectively.
As can be seen, our P2T also achieves superior results compared with recent state-of-the-art 
transformer models.
For example, P2T-Small/Base/Large is 1.1\%/0.5\%/0.6\% better than Swin Transformer \cite{liu2021swin}
with fewer network parameters and less computational cost.
Although PVTv2 \cite{wang2021pvtv2} has large improvement over PVT \cite{wang2021pyramid},
our P2T-Tiny/Small/Base/Large still have 1.1\%/0.4\%/0.3\%/0.3\% improvement 
over PVTv2-B1/B2/B3/B4 \cite{wang2021pvtv2} with fewer parameters and less computational cost.
P2T applies four parallel pooling operations when computing self-attention, still achieving competitive speed with PVTv2 \cite{wang2021pvtv2}.
ViL \cite{zhang2021multi} achieves competitive performance with P2T. 
Nevertheless, the speed of ViL \cite{zhang2021multi} is much slower than our P2T, and the computational cost of ViL \cite{zhang2021multi} is also much larger than P2T.
P2T also largely outperforms ViT \cite{dosovitskiy2021image} and 
DeiT \cite{touvron2021training} with much fewer parameters, 
implying that P2T achieves better performance without a
large amount of training data and knowledge distillation.
Therefore, P2T is very capable for image classification.

\subsection{Semantic Segmentation} \label{sec:exp_sematicseg}
Given a natural image input, semantic segmentation aims at assigning 
a semantic label to each pixel. 
It is one of the most fundamental dense prediction tasks 
in computer vision.

\myPara{Experimental setup.}
We evaluate P2T and its competitors on the ADE20K \cite{zhou2017scene} dataset.
The ADE20K dataset is a challenging scene understanding dataset 
with 150 fine-grained semantic classes. 
This dataset has 20000, 2000, and 3302 images for training,
validation, and testing, respectively.
Following \cite{wang2021pyramid, chu2021twins}, Semantic FPN \cite{kirillov2019panoptic} 
is chosen as the basic method for a fair comparison.
We replace the backbone of Semantic FPN \cite{kirillov2019panoptic}
with various network architectures.
All backbones of semantic FPN have been pretrained on the ImageNet-1K
\cite{russakovsky2015imagenet} dataset, and other layers are initialized 
using the Xavier method \cite{glorot2010understanding}.
All networks are trained for 80k iterations. 
We apply AdamW \cite{loshchilov2017decoupled} as the network optimizer, 
with the initial learning rate of 1e-4 and weight decay of 1e-4.
The \textit{poly} learning rate schedule with $\gamma=0.9$ is adopted.
Each mini-batch has 16 images.
Images are resized and randomly cropped to $512\times 512$ for training.
Synchronized batch normalization across GPUs is also enabled.
During testing, images are resized to the shorter side of 512 pixels.
Multi-scale testing and flipping are disabled.
Following \cite{wang2021pyramid},
we use the MMSegmentation toolbox \cite{mmseg2020} to implement the above experiments.

\myPara{Experimental results.}
Quantitative comparison results are shown in \tabref{tab:exp_ade20k}.
We compare our proposed P2T with ResNets \cite{he2016deep}, 
ResNeXts \cite{xie2017aggregated}, PVT \cite{wang2021pyramid}, 
Swin Transformers \cite{liu2021swin}, Twins \cite{chu2021twins}, and PVTv2 \cite{wang2021pvtv2}.
The results of each network are from the official papers or re-implemented using the official training configurations.
Benefiting from the pyramid pooling technique, the results of 
Semantic FPN \cite{kirillov2019panoptic} with our P2T backbone 
are much better than other CNN and transformer competitors.
Typically, P2T-Tiny/Small/Base/Large are 10.5\%/10.0\%/9.9\%/9.2\% better than ResNet-18/50/101 \cite{he2016deep} and ResNeXt-10-64x4d \cite{xie2017aggregated} with fewer parameters and GFlops, respectively.
Compared with Swin Transformer \cite{liu2021swin} which introduces local self-attention with transformers,
P2T-Small/Base/Large achieve 5.2\%/3.5\%/3.4\% improvement over Swin-T/S/B \cite{liu2021swin}, respectively, 
showing that global relationship modeling is significant for visual recognition.
Twins \cite{chu2021twins} combine the local self-attention from Swin Transformers \cite{liu2021swin} and the global self-attention from PVT \cite{wang2021pyramid}.
As can be observed, Twins \cite{chu2021twins} perform better than Swin Transformers \cite{liu2021swin}, suggesting that global self-attention is significant again.
Unlike Twins \cite{chu2021twins}, we apply pure global self-attention via pyramid pooling, learning richer contexts.
P2T-Small/Base/Large are 3.5\%/3.4\%/2.7\% better than Twins-SVT-S/B/L \cite{chu2021twins}, respectively.
PVTv2 \cite{wang2021pvtv2} is the improved version of PVT \cite{wang2021pyramid}, serving as the strongest competitor for our P2T.
P2T-Tiny/Small/Base/Large are 1.9\%/0.6\%/1.4\%/0.8\% better than PVTv2-B1/B2/B3/B4 \cite{wang2021pvtv2}, respectively.
Besides, P2T-Tiny/Small/Base/Large always have fewer parameters, less computational cost, and faster speed than 
the corresponding PVTv2-B1/B2/B3/B4 \cite{wang2021pvtv2}.
At last, we found that P2T-Tiny is 3.2\% better than ResNeXt-101-64x4d \cite{xie2017aggregated} with twice the speed.
Based on the above observations, we can conclude that P2T is very capable for semantic segmentation.

\subsection{Object Detection} \label{sec:exp_detection}
Object detection is also one of the most fundamental and challenging 
tasks for decades in computer vision.
It aims to detect and recognize instances of semantic objects of certain 
classes in natural images.
Here, we evaluate P2T and its competitors on the MS-COCO \cite{lin2014microsoft} dataset.

\myPara{Experimental setup.}
MS-COCO \cite{lin2014microsoft} is a large-scale challenging dataset 
for object detection, instance segmentation, and keypoint detection.
MS-COCO \texttt{train2017} (118k images) and \texttt{val2017} (5k images) 
sets are used for training and validation in our experiments, respectively.
RetinaNet \cite{lin2017focal} is applied as the basic framework because 
it has been widely acknowledged by this community \cite{liu2021swin, wang2021pyramid}.
Each mini-batch has 16 images with an initial learning rate of 1e-4.
Following the popular MMDetection toolbox \cite{chen2019mmdetection}, 
we train each network for 12 epochs, and the learning rate is divided 
by 10 after 8 and 11 epochs.
The network optimizer is AdamW \cite{loshchilov2017decoupled}, 
a popular optimizer for training transformers.
The weight decay is set as 1e-4.
During training and testing, the shorter side of input images is resized 
to 800 pixels.
The longer side will keep the ratio of the images within 1333 pixels.
In the training stage,
only random horizontal flipping is used for data augmentation.
Standard \texttt{COCO API} is utilized for evaluation, and we report 
results in terms of AP, AP$_{50}$, AP$_{75}$, 
AP$_S$, AP$_M$, and AP$_L$ metrics. 
AP$_S$, AP$_M$, and AP$_L$ mean AP scores for small, medium, and 
large objects defined in \cite{lin2014microsoft}, respectively.
AP is usually viewed as the primary metric.
For each metric, larger scores indicate better performance.
We also report the number of parameters and the computational cost for reference.

\myPara{Experimental results.}
The evaluation results on the MS-COCO dataset are summarized 
in the left part of \tabref{tab:det_seg}.
The results of other networks are from the official papers or re-implemented using the official configurations.
The below discussion refers to the metric of AP if not stated.
We can observe that our P2T achieves the best performance
under all tiny/small/large complexity settings.
For example,
P2T-Small achieves 2.9\%, 1.4\%, and 0.6\% higher AP over Swin-T \cite{liu2021swin}, Twins-SVT-S \cite{chu2021twins}, and PVTv2-B2 \cite{wang2021pvtv2}, respectively.
P2T-Tiny is 1.1\% better than PVTv2 \cite{wang2021pvtv2}.
Compared with ViL \cite{zhang2021multi}, P2T-Tiny/Small are 0.5\% and 0.2\% better than ViL-Tiny/Small \cite{zhang2021multi}, respectively. Note that ViL \cite{zhang2021multi} runs at a much slower speed than P2T, as shown in \tabref{tab:det_seg}.
With the base complexity setting, P2T-Base outperforms Swin-S \cite{wang2021pvtv2} by 1.0\% and is 0.2\% better than the best competitor PVTv2-B3.
With the large complexity setting, P2T-Large achieves 1.1\% and 1.9\% better AP than PVTv2-B4 \cite{wang2021pvtv2} and Twins-SVT-B \cite{chu2021twins}, respectively.
At all complexity levels, P2T always outperforms PVTv2 \cite{wang2021pvtv2} with fewer network parameters, less computational cost, and faster speed.
P2T-Tiny/Small/Base/Large are 9.5\%/8.1\%/7.0\%/6.2\% better than ResNet-18/50/101 \cite{he2016deep} and ResNeXt-101-64x4d \cite{xie2017aggregated}, respectively.
Therefore, P2T is very capable for object detection.

\subsection{Instance Segmentation} \label{sec:exp_instance}
Instance segmentation is another fundamental vision task.
It can be regarded as an advanced case of object detection 
by outputting fine-grained object masks instead of bounding boxes 
in object detection.

\myPara{Experimental setup.}
We evaluate the performance of instance segmentation  
on the well-known MS-COCO dataset \cite{lin2014microsoft}.
MS-COCO \texttt{train2017} and \texttt{val2017} sets are used 
for training and validation in our experiments.
Mask R-CNN \cite{he2020mask} is applied as the basic framework
using different backbone networks.
The training settings are the same as what we use for object detection
in \secref{sec:exp_detection}.
We report evaluation results for object detection and instance 
segmentation in terms of AP$^\text{b}$, AP$_{50}^\text{b}$, AP$_{75}^\text{b}$, 
AP$^\text{m}$, AP$_{50}^\text{m}$, and AP$_{75}^\text{m}$ metrics, 
where ``b'' and ``m'' indicate bounding box and mask metrics, respectively. 
AP$^\text{b}$ and AP$^\text{m}$ are set as the primary evaluation metrics.

\begin{table}[!t]
    \centering
    \setlength{\tabcolsep}{1mm}
    \caption{\textbf{Ablation studies on multiple pyramid pooling ratios.}
    The ``D ratio'' indicates the downsampling ratio between the initial sequence length and the downsampled sequence length.
    ``Top-1'' denotes the top-1 classification accuracy rate on the ImageNet-1K validation set \cite{russakovsky2015imagenet}, and
    ``mIoU'' indicates the results for semantic segmentation on the ADE20K dataset \cite{zhou2017scene}.}
    \label{tab:abla_pool_ratios}
    \resizebox{\columnwidth}{!}{%
    \begin{tabular}{c|ccccc} \Xhline{1pt}
        No. &  Pooling Ratio(s)  & D Ratio $\uparrow$ & Top-1 (\%) $\uparrow$&  mIoU (\%) $\uparrow$ \\ \Xhline{1pt}
        1 & 24 & 576 & 70.6 & 27.5 \\
        2 & 16 & 256 & 72.5 & 33.0 \\
        3 & 12 & 144 & 73.9 & 34.3 \\
        4 & 8 & 64 & 73.9 & 34.4 \\
        5 & 12, 24 & 115 & 74.4 & 34.8\\
        6 & 12, 16, 20, 24 & 66 & 74.7 & 35.7 \\
    \Xhline{1pt}
    \end{tabular}}
\end{table}

\begin{table}[!tb]
\centering
\caption{\textbf{Ablation studies for replacing single pooling operation with multiple pooling operations at different stages.} Since both stage 2 and 3 of our network only have two basic blocks, we merge them into one choice. ``Top-1'' denotes the top-1 classification accuracy rate on the ImageNet-1K validation set \cite{russakovsky2015imagenet}, and
    ``mIoU'' is the results for semantic segmentation on the ADE20K dataset \cite{zhou2017scene}.}
\label{tab:abla_pypool_stages}
\resizebox{\columnwidth}{!}{%
\renewcommand{\tabcolsep}{4mm}
\begin{tabular}{c|ccc|cc} \Xhline{1pt}
\multirow{2}{*}{No.} & \multicolumn{3}{c|}{Stage \# of Network} & \multirow{2}{*}{Top-1 (\%) $\uparrow$} & \multirow{2}{*}{mIoU $\uparrow$}  \\
&  [2, 3] & 4 & 5 & \\
\Xhline{1pt}
1 &  & & & 73.9 & 34.4 \\
2 & \ding{52} & & & 74.1 & 34.9\\
3  & \ding{52} & \ding{52} & & 74.5 & 35.5\\
4  & \ding{52} & \ding{52} & \ding{52} & 74.7 & 35.7\\
\Xhline{1pt}
\end{tabular}}
\end{table}

\begin{table}[!t]
    \centering
    \setlength{\tabcolsep}{4.mm}
    \caption{\textbf{Ablation studies on the choices of pooling operations.}
    We can see that other choices work worse than average pooling.
    ``Top-1'' indicates the top-1 accuracy rate on the ImageNet-1K dataset
    \cite{russakovsky2015imagenet} for image classification. 
    ``mIoU'' is the mean IoU rate on the ADE20K dataset \cite{zhou2017scene} 
    for semantic segmentation.}
    \label{tab:abla_pool_operation}
    \resizebox{.9\columnwidth}{!}{%
    \begin{tabular}{c|cc} \Xhline{1pt}
        Pooling Type & Top-1 (\%) $\uparrow$ &  mIoU (\%) $\uparrow$ \\ \Xhline{1pt}
        Average Pooling  & 74.7 & 35.7 \\
        Max Pooling  & 73.0 & 33.2 \\
        Convolution & 73.8 & 35.5 \\
    \Xhline{1pt}
    \end{tabular}}
\end{table}

\myPara{Experimental results.}
The comparisons between P2T and its competitors are displayed 
in the right part of \tabref{tab:det_seg}.
P2T achieves the best performance
consistently compared to existing CNN and transformer backbone networks.
Compared with transformer-based backbones,
P2T achieves the best performance at all complexity levels.
Regarding the bounding box metric AP$^\text{b}$,
P2T-Small/Base are 3.3\%/2.4\% better than Swin-T/S \cite{liu2021swin}, and
P2T-Small/Large are 2.1\%/3.1\% better than Twins-SVT-S/B \cite{chu2021twins}.
P2T-Tiny/Small/Base/Large are 1.5\%/0.2\%/0.2\%/0.8\% better than PVTv2-B1/B2/B3/B4 \cite{wang2021pvtv2} with fewer parameters, less computational cost, and faster speed, respectively.
In terms of the mask metric AP$^\text{m}$,
we also observe similar improvements as observed using bounding box metrics.
Compared with ResNet-based backbones, P2T significantly outperforms ResNets \cite{he2016deep} and ResNeXts \cite{xie2017aggregated} at all complexity levels.
It is also surprising that our lightest P2T-Tiny is 0.5\% and 1.2\% better than ResNeXt-101-64x4d \cite{xie2017aggregated} in terms of bounding box and mask metrics, respectively.
Therefore, P2T is very capable for instance segmentation.

\subsection{Ablation Studies} \label{sec:ablation}
\myPara{Experimental setup.}
In this section, we perform ablation studies to analyze 
the efficacy of each design choice in P2T.
We evaluate the performance of model variants on semantic segmentation 
and image classification.
Due to the limited computational resources, we only train each model 
variant on the ImageNet dataset \cite{russakovsky2015imagenet} 
for 100 epochs, while other training settings keep the same as 
in \secref{sec:exp_classification}.
Then, we fine-tune the ImageNet-pretrained model on the ADE20K
dataset \cite{zhou2017scene} with the same training settings 
as in \secref{sec:exp_sematicseg}.

\myPara{Multiple pyramid pooling ratios.}
To validate the significance of using multiple pooling ratios,
we conduct experiments to evaluate the performance of P2T with one/two/four parallel pooling operations.
The baseline is P2T-Small without relative positional encoding, IRB, and overlapping patch embedding. 
The results are shown in \tabref{tab:abla_pool_ratios}.
As can be seen, the single pooling operation with a large pooling ratio (\eg, 16, 24) has a large squeezed ratio for the sequence length.
Still, it results in very poor performance for both image classification and semantic segmentation.
However, when the single pooling operation is with a pooling ratio $\le 12$, the performance will be saturated if we further decrease the pooling ratio. When we adopt two parallel pooling operations, even with a high squeezed ratio, the performance still becomes better for both image classification and semantic segmentation. When we have four parallel pooling operations, we can derive the best performance with the comparable squeezed ratio for the pooling ratio of 8 (the setting in PVT \cite{wang2021pyramid}).

\myPara{Significance of pyramid pooling for different stages.
}
We perform ablation studies of the pyramid pooling design of P2T.
for different stages.
Since stage 1 only contains convolutions for downsampling, we do not perform such ablation study at stage 1.
The baseline is same with the last ablation studies. The pooling ratio of single pooling operation is set to 8 for ensuring comparable downsampling ratios.
Results are shown in \tabref{tab:abla_pypool_stages}.
We can observe pyramid pooling can improve the performance at all stages. The performance becomes higher when more stages are applied with multiple pooling operations.
From the results, the improvement on applying multiple pooling operations on stage 4 (No. 3 of \tabref{tab:abla_pypool_stages}) is larger than that on other stages (No. 2, 4 of \tabref{tab:abla_pypool_stages}), because stage 4 has more basic blocks than the summation of stage [2,3] and stage 5.

\begin{table}[!t]
    \centering
    \setlength{\tabcolsep}{0.7mm}
    \caption{\textbf{Ablation study on the fixed pooled size.}
    GFlops is computed with an input size of $512 \times 512$ for 
    the semantic segmentation model, \ie, Semantic FPN 
    \cite{kirillov2019panoptic}.
    Memory (Mem) denotes the training GPU memory usage for Semantic FPN 
    \cite{kirillov2019panoptic} with a batch size of 2.
    ``Top-1'' and ``mIoU'' indicate the top-1 classification accuracy on ImageNet-1K \cite{russakovsky2015imagenet} and segmentation mIoU on ADE20K \cite{zhou2017scene}, respectively.}
    \label{tab:fixed_pooled_size}
    \resizebox{\columnwidth}{!}{%
    \begin{tabular}{c|cccc} \Xhline{1pt}
        Pooling Operation & GFlops $\downarrow$ & Mem (GB) $\downarrow$ & Top-1 (\%) $\uparrow$ &  mIoU (\%) $\uparrow$ \\ \Xhline{1pt}
        Fixed Pooled ratios & 41.6 & 3.3 & 74.7 & 35.7 \\
        Fixed Pooled Sizes & 38.9 & 2.9 & 74.4 & 33.3 \\
    \Xhline{1pt}
    \end{tabular}}
\end{table}

\begin{table}[!t]
    \centering
    \setlength{\tabcolsep}{3.5mm}
    \caption{\textbf{Ablation study on the relative positional encoding (RPE), IRB, and overlapping patch embedding (OPE).}
    ``Top-1'' and ``mIoU'' indicate the top-1 classification accuracy on ImageNet-1K \cite{russakovsky2015imagenet} and segmentation mIoU on ADE20K \cite{zhou2017scene}, respectively.}
    \label{tab:abla_others}
    \resizebox{0.9\columnwidth}{!}{%
    \begin{tabular}{ccc|cc} \Xhline{1pt}
        RPE & IRB & OPE & Top-1 (\%) $\uparrow$ & mIoU (\%) $\uparrow$ \\ \Xhline{1pt}
         &  &  & 74.7 & 35.7 \\
        \ding{52} & & & 76.4 & 37.4 \\
        \ding{52}& \ding{52} & & 79.5 & 42.7 \\
        \ding{52}& \ding{52} & \ding{52} & 79.7 & 44.1 \\
    \Xhline{1pt}
    \end{tabular}}
\end{table}

\myPara{Pooling operation choices.}
We conduct experiments for different pooling operations, as shown in \tabref{tab:abla_pool_operation}.
There are three typical choices, \ie, max pooling, depthwise convolution, and the default average pooling. 
The kernel size of depthwise convolution is the same as max/average pooling to keep the same downsampling rate.
It is obvious that different pooling types do not affect the computational complexity and they only affect the number of the parameters of the downsampling kernel.
Regarding the results of ImageNet classification accuracy \cite{russakovsky2015imagenet} and ADE20K segmentation mIoU \cite{zhou2017scene}, average pooling is much better than the other two choices. Thus, we apply average pooling as the default pooling choice.

\myPara{Fixed pooled sizes.}
When using fixed pooling ratios, the size of the pooled feature map
will vary with the input feature map.
Here, we try to fix the pooled sizes as \{1, 2, 3, 6\} for 
all stages, using adaptive average pooling.
The results are shown in \tabref{tab:fixed_pooled_size}.
Compared with our default setting, 
about 10\% of memory usage and 12\% of computational cost are saved.
However, the top-1 classification accuracy drops by 0.3\%, 
and the semantic segmentation performance is 2.4\% lower. 
Hence, we choose to use fixed pooling ratios rather than fixed pooled sizes.

\myPara{Selection of activation functions.}
We use the Hardswish function \cite{howard2019searching} for nonlinear 
activation to reduce GPU memory usage in the training phase.
Typically, when we train P2T-Small on ImageNet \cite{russakovsky2015imagenet} with a batch size of 64, 
the GPU memory usage of GELU \cite{hendrycks2016gaussian} is 10.5GB, 
which is 3.6GB (+52\%) more than that of Hardswish \cite{howard2019searching}.
We also find that there is no significant accuracy decrease
if we employ Hardswish \cite{howard2019searching}.

\myPara{Other design choices.}
To validate the effectiveness of other design choices like relative positional encoding, IRB, and overlapping patch embedding, we add these components one by one to the baseline.
Experimental results are shown in \tabref{tab:abla_others}.
As can be seen, relative positional encoding has significant improvement for both image classification and semantic segmentation.
With large pooling ratios, the pooled features would have small scales, so relative positional encoding only needs negligible computational cost (5M Flops for the input size of $224\times 224$).
An extra depthwise convolution in the feed-forward network, \ie, IRB,
also shows significant improvement, demonstrating the necessity of 
the 2D locality enhancement.
We further follow \cite{wang2021pvtv2} to add overlapping patch embedding, and 0.2\%/1.4\% improvement is observed for image classification and semantic segmentation, respectively.

\section{Conclusion} %
This paper introduces pyramid pooling into MHSA for alleviating the high computational cost of MHSA in the vision transformer.
Compared with the strategy of applying a single pooling operation in MHSA \cite{wang2021pyramid,fan2021multiscale},
our pooling-based MHSA not only reduces the sequence length but also learns powerful contextual representations simultaneously via pyramid pooling.
Equipped with the pooling-based MHSA, we construct a new backbone network, called Pyramid Pooling Transformer (P2T).
To demonstrate the effectiveness of P2T, we conduct extensive experiments 
on several fundamental vision tasks, including image classification, semantic segmentation,
object detection, and instance segmentation.
Experimental results suggest that P2T significantly outperforms
previous CNN- and transformer-based backbone networks.

\section*{Acknowledgement}

This work is supported in part by Major Project for New 
Generation of AI under Grant No. 2018AAA0100400,
in part by NSFC under Grant No. 61922046, 
in part by Alibaba Innovative Research (AIR) Program,
in part by Alibaba Research Intern Program,
and in part by 
the Agency for Science, Technology and Research (A*STAR) 
under its AME Programmatic Funds (No. A1892b0026 and No. A19E3b0099).

\bibliographystyle{IEEEtran}
\bibliography{reference}

\newcommand{\addSpace}{\vspace{-25pt}}

\newcommand{\AddPhoto}[1]{\includegraphics[width=1in]{#1}}

\addSpace
\begin{IEEEbiography}[\AddPhoto{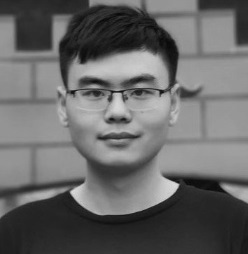}]{Yu-Huan Wu}
 is currently a Ph.D. candidate with the TMCC, College of Computer Science 
 at Nankai University, supervised by Prof. Ming-Ming Cheng. 
 He received his bachelor's degree from Xidian University in 2018. 
 His research interests include computer vision
 and machine learning.
\end{IEEEbiography}

\addSpace
\begin{IEEEbiography}[\AddPhoto{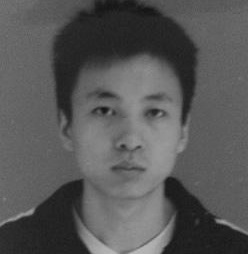}]{Yun Liu}
 received his bachelor's degree and his doctoral degree from 
 Nankai University in 2016 and 2020, respectively.
 His Ph.D. supervisor was Prof. Ming-Ming Cheng.
 Then, he worked with Prof. Luc Van Gool for 
 one and a half years as a postdoctoral scholar 
 at Computer Vision Lab, ETH Zurich.
 Currently, he is a scientist at Institute for Infocomm 
 Research (I2R), A*STAR.
 His research interests include computer vision and 
 machine learning.
\end{IEEEbiography}

\addSpace
\begin{IEEEbiography}[\AddPhoto{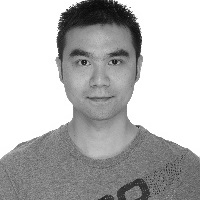}]{Xin Zhan}
 received his bachelor's and doctoral degrees from USTC in 2010 and 2015, respectively.
 Currently, he works as a researcher of Alibaba DAMO Academy.
 His research interests include perception for autonomous driving.
\end{IEEEbiography}

\addSpace
\begin{IEEEbiography}[\AddPhoto{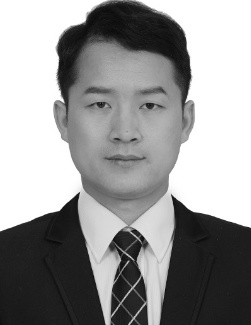}]{Ming-Ming Cheng}
 received his PhD degree from Tsinghua University in 2012.
 Then he did two years research fellow with Prof. Philip Torr
 in Oxford.
 He is now a professor at Nankai University, leading the
 Media Computing Lab.
 His research interests include computer graphics, computer
 vision, and image processing.
 He received research awards, including ACM China Rising Star Award,
 IBM Global SUR Award, and CCF-Intel Young Faculty Researcher Program.
 He is on the editorial boards of IEEE TPAMI/TIP.
\end{IEEEbiography}

\vfill

\end{document}